\documentclass{article}

\PassOptionsToPackage{numbers, compress}{natbib}
\usepackage[final]{tackling_climate_workshop_style}

\usepackage[utf8]{inputenc}
\usepackage[T1]{fontenc}
\usepackage{hyperref}
\usepackage{url}
\usepackage{booktabs}
\usepackage{amsfonts}
\usepackage{amsmath}
\usepackage{nicefrac}
\usepackage{microtype}
\usepackage{graphicx}
\usepackage{placeins}
\usepackage{enumitem}

\title{Beyond Expected Value: PPO with Distributional Critics for Systematic Conservation Planning}

\author{%
  Ana Maria Quintero Ossa \\
  University College London \\
  \texttt{ana.ossa.25@ucl.ac.uk} \\
  \And
  Daniele Silvestro \\
  ETH / CAPTAIN \\
  \texttt{daniele@captaintech.ai} \\
  \And
  James Rudd-Jones \\
  University College London \\
  \texttt{james.rudd-jones.22@ucl.ac.uk} \\
  \And
  Maria Perez-Ortiz \\
  University College London \\
  \texttt{maria.perez@ucl.ac.uk} \\
}

\begin{document}

\maketitle

\begin{abstract}
Systematic conservation planning must allocate limited protection budgets sequentially under ecological and climatic uncertainty. Static tools cannot adapt as conditions change, and policies optimised only for expected outcomes can fail on the worst-case trajectories climate change makes increasingly likely. CAPTAIN was the first framework to cast this as sequential reinforcement learning, but its gradient-free Evolution Strategies (ES) optimiser lacks per-timestep credit assignment. We replace ES with Proximal Policy Optimisation (PPO) via a Plackett-Luce solution to CAPTAIN's non-differentiable action space, and replace its scalar critic with an Implicit Quantile Network (IQN) modelling the full return distribution. A matched-granularity comparison shows decision granularity, not the optimiser, drives most of the ecological gain, so the real contribution is training efficiency and extensibility to risk-sensitive objectives. On a New Zealand marine dataset, it matches or exceeds ES's outcomes at a fraction of the training cost, offering a template for future work ES's black-box design could not support.
\end{abstract}

\section{Introduction} \label{sec:intro}

Biodiversity loss is one of the most consequential and time-critical effects of climate change, and conservation planning to slow it is fundamentally a sequential decision problem \citep{margules2000systematic}: which sites to protect, in what order, under a fixed and shrinking budget, while species distributions, disturbance regimes, and climate trajectories keep changing. Static prioritisation tools such as Marxan \citep{ball2009marxan} and Zonation \citep{lehtomaki2013methods} compute one allocation for a fixed snapshot of the data and have no native mechanism for adapting that allocation over time. CAPTAIN (Conservation Area Prioritisation Through Artificial INtelligence) \citep{silvestro2022improving} was the first framework to cast this problem in a form that could be explored with Reinforcement Learning (RL), and showed a learned policy can substantially outperform Marxan. Its original optimiser, Parallelised Evolution Strategies (ES) \citep{salimans2017evolution}, is gradient-free and evaluates whole episodes. It updates parameters proportional to the resulting return, with no mechanism to attribute that return to the specific decisions that produced it, and no representation beyond the expected outcome.

This second limitation matters specifically for climate applications. A policy optimised purely for expected biodiversity outcomes can still perform badly on the worst-case climate trajectories that conservation practitioners most need to plan for. Addressing this requires more than a better point estimate: it requires a policy that represents how outcomes vary across possible futures, so a risk-sensitive objective has something to act on. Distributional RL, which learns the full distribution of returns rather than only its mean \citep{bellemare2017distributional,dabney2018implicit}, offers a principled route to exactly this, and from there to a risk-sensitive objective such as Conditional Value-at-Risk (CVaR) \citep{rockafellar2000optimization}. Whether this combination, gradient-based credit assignment plus a distributional, risk-aware critic, can be made to work inside an existing, large-scale conservation-planning framework had not been tested. PPO is a natural candidate for this setting: it pairs stable, sample-efficient policy-gradient updates \citep{schulman2017proximal} with Generalised Advantage Estimation for genuine per-timestep credit assignment \citep{schulman2016highdimensional}, directly addressing the credit-assignment limitation of ES described above. A distributional critic is a natural complement: beyond enabling risk-sensitive objectives, learning the full return distribution has been shown to improve policy quality even without an explicit risk-sensitive objective \citep{bellemare2017distributional}, suggesting it may help here independently of whether CVaR itself proves useful.

We make three contributions:
\begin{itemize}[leftmargin=*,noitemsep,topsep=2pt]
\item We replace CAPTAIN's ES optimiser with PPO \citep{schulman2017proximal}, solving CAPTAIN's non-differentiable top-$K$ cell-selection action space via Plackett-Luce sequential sampling, and redesign its reward structure for genuine per-timestep credit assignment; isolating this from a decision-granularity confound via a matched-granularity ES comparison shows granularity, not the optimiser, is the primary driver of the resulting Endangered-category recovery, and that PPO's genuine contribution is training efficiency and extensibility to a distributional critic (Section~\ref{sec:method}, \ref{sec:results}).
\item We replace PPO's scalar critic with an Implicit Quantile Network (IQN) learning the full return distribution, and show it improves both conservation outcomes and training efficiency over PPO alone, with no risk-sensitive objective required (Section~\ref{sec:results}).
\item We additionally test a CVaR-based risk-sensitive objective on top of the IQN critic, and report an honest boundary-condition finding: it does not yet help, because the current disturbance regime does not generate enough outcome variance for a tail-risk objective to act on (Section~\ref{sec:results}).
\end{itemize}

\section{Method} \label{sec:method}

We build on CAPTAIN v3\footnote{\url{https://github.com/captain-project/captain3preview}}, evaluated here on a New Zealand marine biodiversity dataset \citep{stephenson2023atlas,guevara2025use} (583 species, a 17{,}000-cell protection budget). At each timestep, the agent selects cells for protection from a grid of candidate sites, guided by Species Distribution Models (SDMs) that project habitat suitability under a climate trajectory and by each species' IUCN Red List risk category.\footnote{Least Concern (LC), Near Threatened (NT), Vulnerable (VU), Endangered (EN), Critically Endangered (CR).} Appendix~\ref{sec:dataset} gives the full input-layer breakdown. The category counts we report throughout (e.g.\ Table~\ref{tab:results}) are the model's internally projected risk level under its own habitat-suitability model and climate trajectory -- a simulated estimate of protection's ecological effect, not an empirical claim about how quickly real-world IUCN assessments are revised.

\paragraph{ES baseline.} CAPTAIN's original optimiser \citep{salimans2017evolution} treats the policy network as a black box. Each iteration, 32 perturbed copies of the policy parameters are evaluated across parallel workers by running full episodes, and the parameters are updated in the direction of a fitness-weighted average of the perturbations, with greedy top-$K$ cell selection ($K{=}1{,}000$ cells per timestep) and no gradient computation anywhere in the loop. This coarse granularity suits ES specifically: with no per-timestep signal to exploit either way, spending the budget in fewer, larger steps is cheaper to simulate across a large population of full-episode rollouts. It is nonetheless a genuine limitation for credit assignment, not merely a stylistic one: with no per-timestep signal, ES cannot distinguish an early protection decision that mattered from one that did not, and it also has no differentiable action-selection probability for a policy-gradient method to optimise against in the first place.

\paragraph{PPO via Plackett-Luce sampling.} We replace ES with PPO \citep{schulman2017proximal}, using Generalised Advantage Estimation \citep{schulman2016highdimensional} for genuine per-timestep credit assignment. CAPTAIN's original architecture scores each cell independently and selects the top-$K$, with no associated probability distribution to differentiate. We resolve this via Plackett-Luce sequential sampling \citep{plackett1975analysis,luce1959individual,yadav2021policygradient}: the $K$ selections are drawn sequentially without replacement, each cell sampled with probability proportional to $\mathrm{softmax}$ over the remaining candidates' scores, giving an exact, differentiable log-probability $\log \pi(a \mid s) = \sum_{i=1}^{K} \log \big(\exp(\mathrm{score}_i) / \sum_{j \in \mathrm{remaining}} \exp(\mathrm{score}_j)\big)$, deterministic (greedy top-$K$, no sampling noise) at evaluation time. This sum makes ES's $K{=}1{,}000$ impractical for PPO regardless: the log-probability grows highly negative, destabilising the surrogate objective. We instead use $K{=}50$ over 340 timesteps (same total budget, finer-grained decision sequence). Additionally, re-train ES at this same $K{=}50$ to isolate whether outcome differences reflect the optimiser or simply the finer decision sequence (Section~\ref{sec:results}). CAPTAIN's original reward, a delta-based extinction-risk signal plus a cumulative cost penalty, suits ES's whole-episode evaluation but leaves PPO's critic absorbing the timestep-varying expected cell value, collapsing advantages to near zero. We redesign it around a counterfactual (regret-based) baseline: at each step we compare newly selected cells' ecological value against the still-available pool's mean. Hence, a random policy scores approximately zero by construction. A separate hard constraint enforces the total protection budget, not the reward.

\paragraph{Stochastic disturbance.} To create genuine environmental stochasticity, distinct from the policy's own randomness, we add a resampled Perlin-noise disturbance layer: at the start of each episode, a disturbance intensity is drawn from $\mathrm{Uniform}(0.05, 0.5)$, producing a spatially coherent field of disturbance events that varies episode to episode. This is intended to represent uncertainty in future disturbance severity, e.g.\ marine heatwave frequency, that a real deployment would face. It is a synthetic noise pattern, though, not something derived from an actual climate model: it captures spatial coherence and variation from episode to episode, but not the physical drivers or longer-term trends behind real disturbance events, a simplification future work should address directly.

\paragraph{Distributional critic and CVaR filtering.} We replace PPO's scalar critic with an Implicit Quantile Network (IQN) \citep{dabney2018implicit}, which learns the return distribution $Z(s)$ directly by parameterising its quantile function via a learned embedding of the quantile level $\tau$, trained with a quantile Huber loss over $n{=}8$ sampled quantiles per update. We additionally test a CVaR objective \citep{rockafellar2000optimization} built on this critic, filtering at the episode level rather than inside the Bellman recursion (known to be time-inconsistent under naive application \citep{bauerle2011markov}): we retain only the worst $\alpha{=}0.25$ fraction of episodes by ecological outcome and train PPO on those alone, against an $\alpha{=}1$ (no filtering) control isolating the critic's own contribution. By construction, CVaR$_\alpha$ is the expectation over the worst $\alpha$ fraction of outcomes, not the single worst case; we chose $\alpha{=}0.25$ as a standard tail-focused setting that still retains enough episodes per update to estimate the objective reliably, with tighter $\alpha$ a natural direction for future work.

\FloatBarrier
\section{Experiments and Results} \label{sec:results}

\begin{table}[t]
\caption{final species counts by IUCN category (583 species total; greedy top-$K$ evaluation, seed-varied over 10 disturbance draws; zero outcome variance across seeds for every model). Time/iter is wall-clock seconds per training iteration (ES epoch or PPO/IQN update); Total is the full training run's wall-clock cost to the checkpoint reported here (computed from logged per-iteration time $\times$ iteration count), not a per-iteration unit, since an ES epoch and a PPO/IQN update are different units of work. Same machine throughout (specifications in Appendix~\ref{sec:compute}).}
\label{tab:results}
\centering
\small
\begin{tabular}{lcccccccc}
\toprule
\textbf{Model} & \textbf{LC} & \textbf{NT} & \textbf{VU} & \textbf{EN} & \textbf{CR} & \textbf{Time/iter} & \textbf{Iters} & \textbf{Total} \\
\midrule
No protection                          & 395 & 86 & 30 & 12 & 60 & -- & -- & -- \\
ES, stochastic disturbance ($K{=}1{,}000$) & 532 & 24 & 18 &  4 &  5 & $\sim$300s & -- & -- \\
\textbf{ES, stochastic disturbance ($K{=}50$)}  & \textbf{545} & \textbf{27} & \textbf{6} & \textbf{0} & \textbf{5} & \textbf{$\sim$1900s} & \textbf{110} & \textbf{$\sim$58h} \\
\textbf{PPO, stochastic disturbance ($K{=}50$)} & \textbf{530} & \textbf{38} & \textbf{9} & \textbf{0} & \textbf{6} & \textbf{$\sim$227s} & \textbf{300} & \textbf{$\sim$19h} \\
\textbf{IQN, $\alpha=1$ ($K{=}50$)}             & \textbf{540} & \textbf{28} & \textbf{10} & \textbf{0} & \textbf{5} & \textbf{$\sim$770s} & \textbf{200} & \textbf{$\sim$43h} \\
IQN, $\alpha=0.25$ (CVaR, $K{=}50$)     & 548 & 20 &  8 &  2 &  5 & $\sim$770s & 200 & $\sim$55h \\
\bottomrule
\end{tabular}
\end{table}

Table~\ref{tab:results} reports final species counts by IUCN category under greedy top-$K$ evaluation, seed-varied over 10 independent disturbance draws. Every model shows zero outcome variance across those seeds, a result we return to below. Each row reflects a single training run; given how small the margins are between the strongest variants, these comparisons should be read as indicative rather than statistically conclusive, and multi-seed replication is left to future work.

\begin{figure}[h]
\centering
\includegraphics[width=\linewidth]{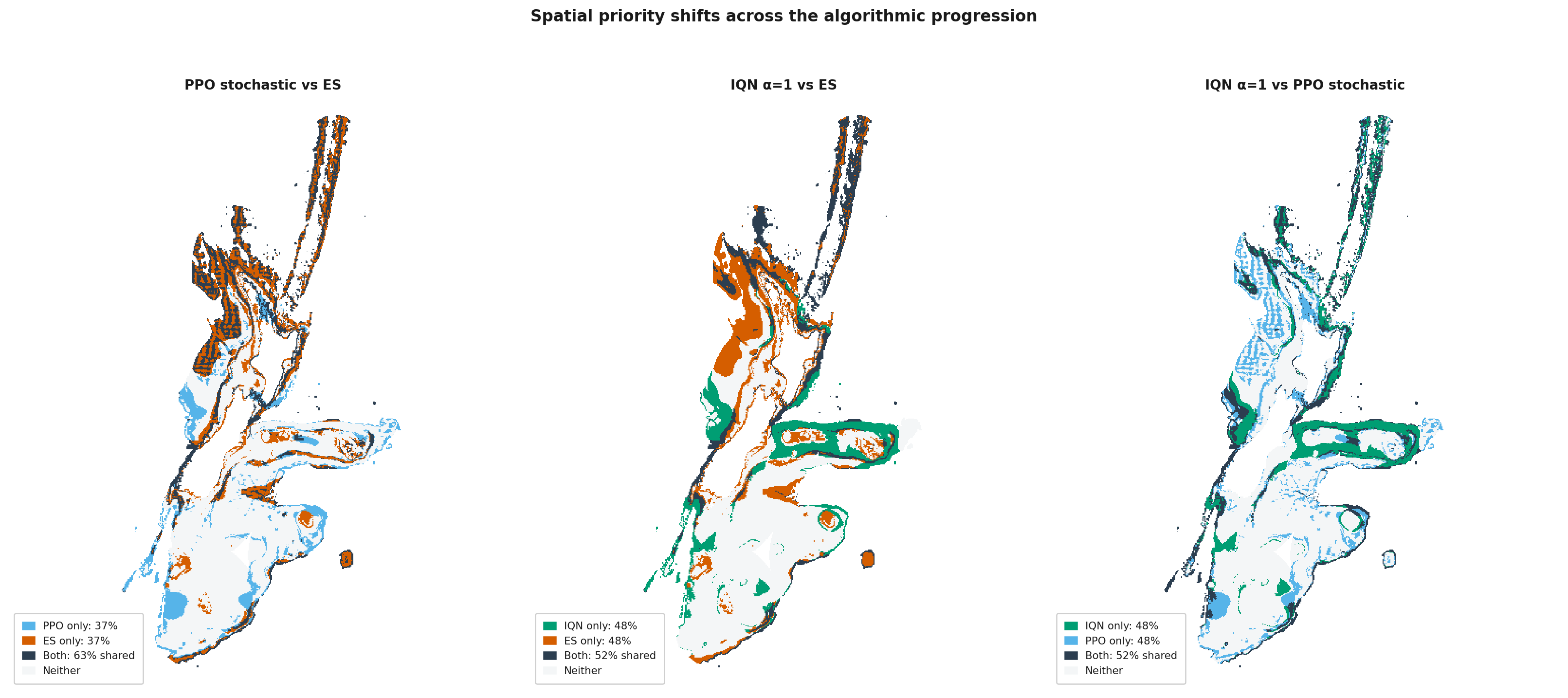}
\caption{Spatial priority shifts across the algorithmic progression (ES refers to the $K{=}50$ stochastic-disturbance variant throughout). Each panel shows which protected cells are unique to one method vs.\ shared with the other, for each consecutive pair (ES$\to$PPO, ES$\to$IQN, PPO$\to$IQN). Overlap and unique percentages are labelled in each panel's legend.}
\label{fig:spatial}
\end{figure}

\paragraph{Decision granularity, not the optimiser, drives the Endangered recovery.} ES at its original $K{=}1{,}000$ granularity leaves species Endangered that PPO at $K{=}50$ clears entirely (Table~\ref{tab:results}). To test whether this reflects PPO's credit assignment or simply its finer-grained decision sequence, we re-trained ES at matched $K{=}50$ granularity: it also clears Endangered, and matches or beats every PPO/IQN variant except the CVaR-filtered run on Vulnerable and Near Threatened too. Granularity, not the optimiser, is therefore the primary driver of the recovery: our original ES-$K{=}1{,}000$-vs-PPO comparison overstated PPO's algorithmic contribution. PPO's genuine advantage over ES at matched granularity is training efficiency, not raw outcome quality. A per-iteration comparison understates this, since an ES epoch (64 perturbation rollouts) and a PPO update (one trajectory) are different units of work: comparing total wall-clock cost to the checkpoints reported in Table~\ref{tab:results} instead, PPO's full run totals $\sim$19h against ES $K{=}50$'s $\sim$58h -- roughly $3\times$ cheaper in total, not merely per update. Figure~\ref{fig:spatial} (Appendix~\ref{sec:spatial-figure}) confirms this spatially: PPO's allocation remains majority-shared with ES $K{=}50$'s, despite an identical budget.

\begin{figure}[h]
\centering
\includegraphics[width=0.55\linewidth]{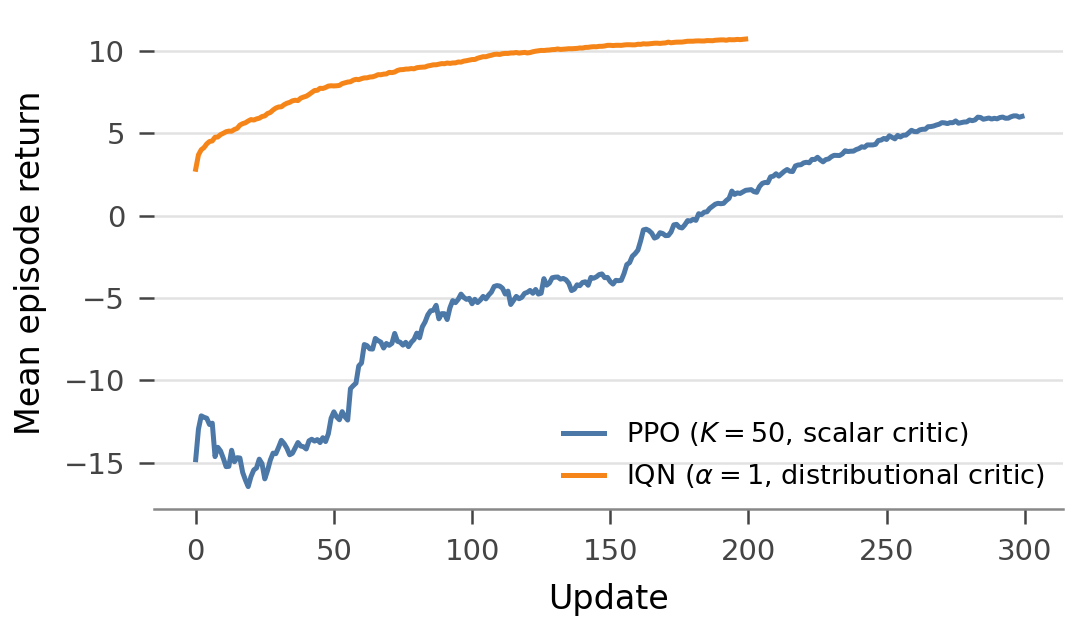}
\caption{Mean episode return over training for the PPO-$K{=}50$ and IQN $\alpha{=}1$ runs reported in Table~\ref{tab:results}. IQN plateaus by roughly update 150; PPO is still improving at update 300 and does not reach IQN's level within its training budget.}
\label{fig:training}
\end{figure}

\paragraph{Scalar vs.\ distributional critic.} Swapping PPO's scalar critic for the IQN distributional critic improves outcomes further, without any CVaR filtering: the $\alpha{=}1$ control matches PPO on Endangered while cutting Near Threatened further, and converges markedly faster: its return plateaus by update $\sim$150, while PPO's is still rising at update 300 without reaching IQN's level (Figure~\ref{fig:training}, Appendix~\ref{sec:training-curves}), despite a higher per-update cost (Table~\ref{tab:results}). This holds under a total-cost framing too: IQN reaches its plateau at a total cost of $\sim$32h (its $\sim$150-update convergence point), and PPO has not matched that return level even after its own full $\sim$19h, 300-update budget -- whether PPO would eventually catch up given a larger budget is untested, but within the budgets we trained both at, IQN's higher per-update cost buys a return PPO does not reach. Notably, IQN differs from both PPO and ES $K{=}50$ by the same substantial margin (48\%), while PPO and ES $K{=}50$ remain majority-shared (Figure~\ref{fig:spatial}, Appendix~\ref{sec:spatial-figure}) -- the distributional critic reshapes the spatial allocation more than the optimiser switch does. We read this as a two-tier effect: all three methods agree on the small, high-penalty EN/CR habitat that dominates the reward signal, but PPO and IQN diverge on the larger, more ambiguous NT/VU cells. At the margin, critic architecture matters more than optimiser choice.

\paragraph{CVaR filtering: a boundary-condition finding.} CVaR filtering ($\alpha{=}0.25$) does not improve on the $\alpha{=}1$ control on any metric (Table~\ref{tab:results}).\footnote{A training-volume confound also affects this comparison: the $\alpha{=}0.25$ run used twice as many episodes per update as the $\alpha{=}1$ control.} With zero outcome variance across 10 disturbance seeds for every model, the current single-scenario, mild-disturbance environment produces no genuine worst-case tail for CVaR to target -- a boundary condition rather than a failure. Two compounding factors drive this: the climate trajectory is a single interpolated scenario fixed per episode, contributing no return variance of its own; and the sampled disturbance intensity range ($[0.05, 0.5]$) is mild enough that none of the resulting disturbance fields shift any species across an IUCN category boundary differently from any other under greedy top-$K$ evaluation. The resulting return distribution is narrow enough that its worst 25\% looks like its average, leaving CVaR filtering nothing to act on -- discussed further in Section~\ref{sec:pathway}.

\section{Pathway to Climate Impact} \label{sec:pathway}

This work targets a concrete deployment gap in climate-driven conservation planning and directly supports SDG 13 (Climate Action), SDG 14 (Life Below Water), and SDG 15 (Life on Land). Every RL-for-conservation method we are aware of \citep{lapeyrolerie2022deep,equihua2024connectivity} optimises expected outcomes only, and none has been validated under the kind of environmental disturbance a changing climate actually produces. Our pipeline addresses this directly, operating under per-episode disturbance variation rather than a single fixed scenario and extending to a distributional, risk-aware critic that ES cannot support, at a fraction of ES's training cost (Table~\ref{tab:results})—itself climate-relevant, given the compute cost of training. A CVaR-based risk-sensitive objective layered on top does not yet help (Section~\ref{sec:results}), so a deployment with only one climate scenario or mild disturbance should invest in representing multiple scenarios or a wider disturbance range before relying on it.

We also note a limit on how directly the model's output can be acted on: translating a recommended allocation into actual protected status involves legal, political, and stakeholder processes the model does not represent. CAPTAIN-style tools are intended to inform prioritisation and sequencing for human decision-makers, not to execute protection decisions unilaterally. Relatedly, the episode's 340 timesteps are sequential decision points sized to the protection budget (17{,}000 cells at $K{=}50$ per step), not a calendar-dated trajectory: the present/future SDM pair spans one interpolated climate scenario rather than a scenario pinned to a specific real-world year, so no fixed real-world duration is currently attached to an episode. Attaching one, by tying timesteps to the underlying climate projection's own horizon, is a concrete gap future deployment work should close before treating any of these counts as timed predictions.

The pipeline itself is SDM-agnostic and location-independent by construction: it consumes standard species distribution model outputs and IUCN risk categories, not anything specific to the present marine environment, and the same code operates on any comparably structured dataset, marine or terrestrial. We see this cross-domain portability, rather than performance on any one dataset, as the paper's main pathway to real-world climate impact.

\section{Conclusion} \label{sec:conclusion}

We present a PPO-and-Plackett-Luce reformulation of CAPTAIN that, unlike its original gradient-free ES optimiser, supports a distributional (IQN) critic and achieves the lowest Near Threatened count of any method tested. We see this extensible, gradient-based formulation as the paper's main contribution: a template for future risk-sensitive conservation planning, whether via stronger CVaR objectives or richer, multi-scenario SDM inputs. PPO's genuine advantage over an equally granular ES is training cost, not raw outcome quality (Section~\ref{sec:results}). CVaR-based episode filtering, by contrast, remains a null result: it needs environmental stochasticity that the present single-scenario setup does not yet provide. Future work will test the pipeline against multiple climate scenarios and a wider disturbance range, where we expect CVaR filtering to become meaningful.

\begin{ack}
The authors thank the CAPTAIN development team for making the framework and codebase available, and for guidance on reward design.
\end{ack}

\bibliographystyle{plainnat}
\bibliography{references}

\clearpage
\appendix
\section{Spatial Allocation Figure} \label{sec:spatial-figure}

\section{Training Curves} \label{sec:training-curves}

\section{Compute} \label{sec:compute}

All models (ES, PPO, and IQN, including both the original $K{=}1{,}000$ ES runs and the matched-granularity $K{=}50$ comparison) were trained on the same machine: a single NVIDIA GeForce RTX 2080 Ti GPU (11\,GB VRAM), an Intel Core i7-9700K CPU (8 cores, 3.60\,GHz), and 32\,GB RAM. No model in Table~\ref{tab:results} ran on CPU only; the reported Time/iter figures are therefore directly comparable.

\section{Dataset} \label{sec:dataset}

The environment is an $800{\times}500$-cell spatial grid. Each of the 583 species has a paired present- and future-condition species distribution model raster, providing per-cell, per-species habitat suitability under the current climate and one interpolated future scenario; a per-species trait table records conservation status, disturbance sensitivity, dispersal ability, and growth rate. Shared spatial layers include protection cost (present and future), a disturbance-exposure layer (present and future, the basis for the stochastic disturbance described in Section~\ref{sec:method}), and present/future sea temperature. All layers share the same grid resolution and are stacked per-cell to form the agent's observation.

\section{Hyperparameters}

\begin{table}[h]
\caption{ES baseline configuration ($K{=}1{,}000$; the $K{=}50$ comparison run uses $K{=}50$, 340 timesteps, 64 perturbations, 8 parallel workers, evaluated at epoch 110/150, otherwise unchanged).}
\centering
\small
\begin{tabular}{ll}
\toprule
Parameter & Value \\
\midrule
Cells per step ($K$) & 1{,}000 \\
Timesteps per episode & 50 \\
Perturbations & 32 \\
Parallel workers & 4 \\
Initial $\sigma$ & 0.3 \\
Initial $\alpha$ & 0.2 \\
Target protected cells & 17{,}000 \\
\bottomrule
\end{tabular}
\end{table}

\begin{table}[h]
\caption{PPO and IQN configuration (shared unless noted).}
\centering
\small
\begin{tabular}{ll}
\toprule
Parameter & Value \\
\midrule
Cells per step ($K$) & 50 \\
Timesteps per episode & 340 \\
Discount $\gamma$ & 0.99 \\
GAE $\lambda$ & 0.95 \\
PPO clip coefficient & 0.2 \\
Learning rate & $3\times10^{-4}$ (annealed) \\
Entropy coefficient & 0.02 \\
Hidden dimensions & [64, 64] \\
Value-function coefficient & 0.25 \\
Disturbance intensity range & $[0.05, 0.5]$ \\
Disturbance coherence & 8 \\
Disturbance impact factor & 0.5 \\
IQN quantile samples & 8 \\
IQN cosine embedding dim.\ & 64 \\
CVaR $\alpha$ (filtered run) & 0.25 \\
Episodes per update ($\alpha{=}0.25$) & 20 \\
Episodes per update ($\alpha{=}1$ control) & 10 \\
Updates (PPO, no distributional critic) & 300 \\
Updates (IQN, both $\alpha$ variants) & 200 \\
\bottomrule
\end{tabular}
\end{table}

\end{document}